\documentclass[10pt,twocolumn,letterpaper]{article}

\usepackage{cvpr}              
\usepackage{bbding}
\usepackage{multirow}
\usepackage{booktabs}
\usepackage{algorithm}
\usepackage{algorithmic}
\usepackage{tabularx, booktabs, array, multirow}
\newcolumntype{Y}{>{\centering\arraybackslash}X}   
\newcolumntype{Z}{>{\raggedright\arraybackslash}X} 

\usepackage{caption} 

\definecolor{cvprblue}{rgb}{0.21,0.49,0.74}
\usepackage[pagebackref,breaklinks,colorlinks,allcolors=cvprblue]{hyperref}

\def\paperID{446} 
\def\confName{CVPR}
\def\confYear{2026}

\title{DarkShake-DVS: Event-based Human Action Recognition under Low-light and Shaking Camera Conditions}

\author{
    Jiaqi Chen\textsuperscript{\rm 1},\space 
    Qinfu Xu \textsuperscript{\rm 1},\space
    Liyuan Pan\textsuperscript{\rm 1}\thanks{Corresponding author} \space
   \\
    \textsuperscript{1}Beijing Institute of Technology \space
    \\
    {\tt\small jq\_chen924@163.com, liyuan.pan@bit.edu.cn}
}
\begin{document}
\maketitle

\begin{abstract}

Human Action Recognition (HAR) is a fundamental computer vision task with diverse real-world applications. Practical deployments often involve low-light environments and unconstrained 6-DoF camera motion, conditions that degrade visual quality, disrupt temporal coherence, and compromise reliability of existing methods. Event cameras, with high low-light sensitivity and microsecond-level temporal resolution, paired with an inertial measurement unit (IMU), present a promising solution. However, current research faces two key challenges: absence of a benchmark integrating low-light conditions, 6-DoF motion, and synchronized IMU data; and lack of effective motion compensation techniques.
To address these, we propose Event–IMU Stabilized HAR (EIS-HAR), with two modules. The first is an EIS module that reduces motion blur via a non-linear warping function to reconstruct a motion-compensated input. The second is a HAR module with a four-stage hybrid architecture to efficiently extract spatiotemporal features for accurate action recognition. To alleviate data scarcity, we introduce DarkShake-DVS, the first large-scale event-based HAR benchmark that includes 18,041 real-world clips captured in low light and intense 6-DoF motion, supplemented by synchronized IMU data. Extensive experiments on three datasets demonstrate consistent superiority of EIS-HAR over state-of-the-art methods. \href{https://github.com/Typistchen/DarkShake-DVS}{[Code]}

\vspace{-5mm}
\end{abstract}

\section{Introduction}
\label{sec:intro}

Human Action Recognition (HAR) is a core task in computer vision with broad applications in nighttime monitoring \cite{liustoryboard, yang2024event}, handheld devices \cite{7917345, 10812874}, and drone-mounted platforms \cite{10734189, 10777515}. Most existing methods rely on idealized assumptions of sufficient illumination and stationary cameras \cite{Xie_2025_CVPR, Zhu_2025_CVPR}. However, these assumptions are frequently violated in real-world deployments, which often involve low-light environments and unconstrained six degrees-of-freedom (6-DoF) camera motion.

Such conditions pose significant challenges: low light reduces the signal-to-noise ratio, while camera motion introduces motion blur, collectively degrading spatial appearance and temporal coherence. Despite their prevalence, these factors have been largely studied in isolation—prior work typically addresses either low-light scenarios \cite{wang2024dailydvs} or camera motion \cite{zhao2023event}—resulting in substantial performance degradation when both are present.

\begin{figure}
    \centering
  \includegraphics[scale=0.65]{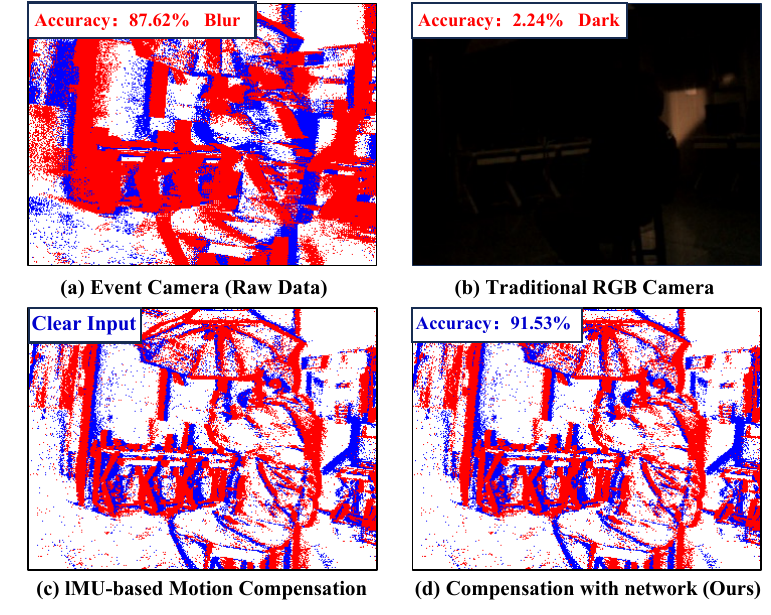}
    \caption{ 
    {\textit{RGB limitations and the effectiveness of IMU-based compensation for event streams in low light with 6-DoF camera motion. (a) Raw event data suffering from  motion-induced distortions, achieving a baseline accuracy of 87.62\%. (b) The conventional RGB camera at the same location cannot provide reliable visual information. (c) Our IMU-based motion compensation method successfully stabilizes the event stream, generating a clear, structured input. (d) By processing this compensated data, our network achieves a significantly improved accuracy of 91.53\%.}} 
    }

    \label{fig:intro}
\vspace{-3mm}
\end{figure}

\begin{figure*}
    \centering
  \includegraphics[scale=0.5]{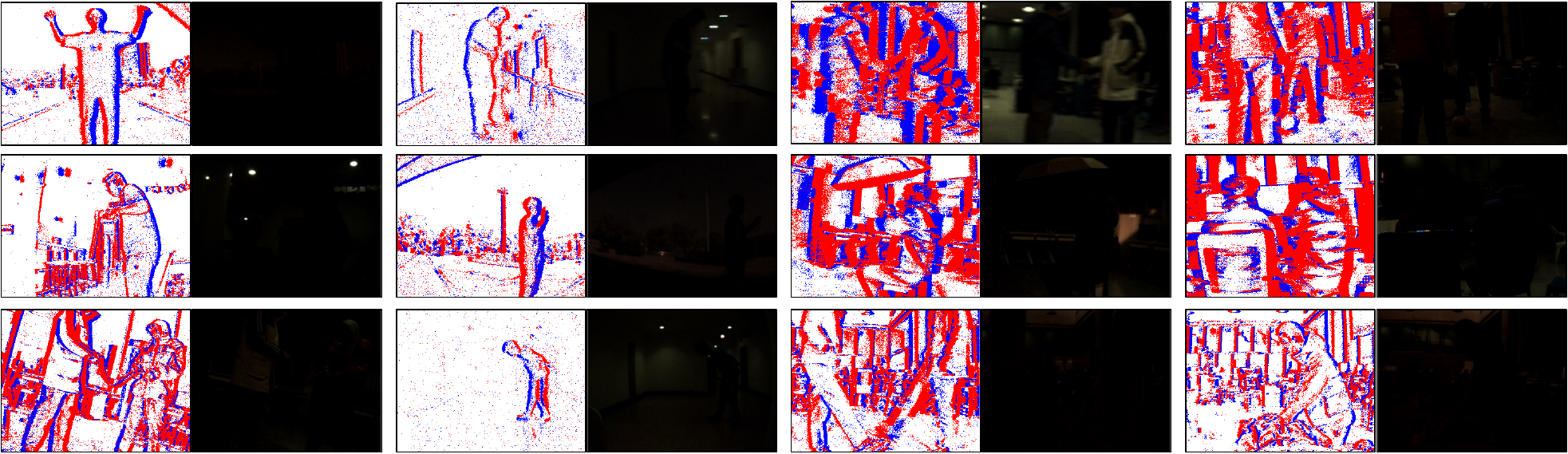}
    \caption{ \textit{\textbf{Examples of our DarkShake-DVS dataset}.
It contains 18K pairs of  RGB frames and event streams, covering both indoor and outdoor scenes.
The examples show data captured under different scenarios, as well as samples with varying degrees of camera motion.}
    } 
    \label{fig:dataset}
   \vspace{-3mm}
\end{figure*}

Under these conditions, conventional RGB sensors struggle to provide reliable visual information (Fig.~\ref{fig:intro}(b)), which in turn impairs the performance of RGB-based methods \cite{Xie_2025_CVPR, pan2020high, pan2020single}. To address these limitations, we explore bio-inspired event cameras \cite{6889103, 4444573}, which offer low-light sensitivity \cite{10.1007/978-3-031-72775-7_17, Yang_2023_ICCV} and microsecond temporal resolution \cite{Pan_2019_CVPR, Yang_2025_CVPR} for motion scenarios. Furthermore, IMU data synchronized with event streams provide angular velocity and linear acceleration, supplying motion cues for ego-motion estimation and stabilizing event streams under 6-DoF motion. 

Despite these advantages, current event-based HAR methods \cite{wang2024hardvs, liu2022video, zhouspikformer, li2024videomamba} perform poorly under combined low-light and 6-DoF camera motion. We identify two key reasons: first, benchmarks lack data that jointly capture low-light conditions, ego-motion, and IMU data; second, existing event-based HAR methods rarely incorporate explicit motion compensation, and IMU-based methods remain largely unexplored. Thus, a dedicated benchmark for this joint setting with synchronized IMU, together with motion compensation methods, is needed to advance research.


To this end, we propose Event–IMU Stabilized Human Action Recognition (EIS-HAR), a comprehensive framework. First, we introduce DarkShake-DVS, the first large-scale event-based HAR benchmark recorded under low light and severe 6-DoF ego-motion, with high-frequency IMU data. The dataset includes 18K real-world clips, covering 30 single-person actions and 32 multi-person interactions. Second, to mitigate distortions, we design Adaptive IMU-based Motion Compensation (AIMC), which parses IMU data to dynamically adjust temporal processing windows and employs a non-linear warping function to efficiently reconstruct motion-compensated event streams (Fig.~\ref{fig:intro}(c)), providing clean input for recognition. Finally, to validate the dataset’s challenge and compensation efficacy, we propose a robust recognition framework. Due to increased temporal density after compensation, we first apply Iterative Greedy Sampling (IGS)—which uses dynamic suppression weighted by relevance, quality, uniformity, and diversity—to select informative key frames. Building on these, we develop the Hybrid Spatio-Temporal Swin Transformer (HSTS), a four-stage hybrid architecture that jointly captures long-range structure and local spatiotemporal cues. 

Extensive experiments on HARDVS, DailyDVS-200, and DarkShake-DVS demonstrate that our framework outperforms state-of-the-art methods, validating robustness under these joint extreme conditions. 

Our main contributions are summarized as follows:

\begin{itemize}

\item We build DarkShake-DVS, the first large-scale event-based HAR benchmark that explicitly couples low illumination with authentic 6-DoF ego-motion.

\item We propose an efficient IMU-based adaptive motion compensation framework capable of high-fidelity removal of spatio-temporal distortions caused by camera shake.

\item We introduce an Iterative Greedy Sampling scheme and a Hybrid Spatio-Temporal Swin model, which together form a framework for stabilized event-based HAR.

\end{itemize}

\section{Related Work}

\textbf{Event Benchmark Datasets for HAR.} As shown in Tab. \ref{table1:dataset}, event-based action recognition datasets have evolved from artificial data \cite{orchard2015converting} to complex real-world scenes \cite{amir2017low, gao2023action, wang2024hardvs, wang2024event}.
However, existing datasets share limitations: they lack data captured  under low illumination with 6-DoF ego-
motion. This hinders the evaluation of event camera robustness and stability in extreme conditions.
To fill this gap, we introduce DarkShake-DVS, the first action dataset featuring both darkness and severe shake, designed to benchmark event-driven action recognition in challenging scenarios.

\begin{table*}[]

\scalebox{0.8}{
\begin{tabular}{ccccccccccccc}
\toprule
\textbf{Dataset}      & \textbf{Year} & \textbf{Sensors} & \textbf{Scale} & \textbf{Class} & \textbf{Resolution} & \textbf{Real}                 & \textbf{Dark}                 & \textbf{Shake}                & \textbf{M-P}                  & \textbf{M-VW/ILL/MO}          & \textbf{DYB}                  & \textbf{OCC}                  \\ \hline
ASLAN-DVS    & 2011          & DAVIS240c        & 3,697          & 432            & 240×180             & \XSolidBrush                              & -                             & -                             & -                             & -                             & -                             & -                              \\
MNIST-DVS    & 2013          & DAVIS128         & 30,000         & 10             & 128×128             & \XSolidBrush                              & -                             & -                             & -                             & -                             & -                             &  -                             \\
N-Caltech101 & 2015          & ATIS             & 8,709          & 101            & 302×245             & \XSolidBrush                              & -                             & -                             & -                             & -                             & -                             & -                              \\
N-MNIST      & 2015          & ATIS             & 70,000         & 10             & 28×28               & \XSolidBrush                              & -                             & -                             & -                             & -                             & -                             & -                              \\
CIFAR10-DVS  & 2017          & ATISDAVIS128     & 10,000         & 10             & 128×128             & \XSolidBrush                              & -                             & -                             & -                             & -                             & -                             & -                              \\
HMDB-DVS     & 2019          & DAVIS240c        & 6,766          & 51             & 240×180             &   \XSolidBrush                            & -                             & -                             & -                             & -                             & -                             & -                              \\
UCF-DVS     & 2019          & DAVIS240c        & 13,320         & 101            & 240×180             &  \XSolidBrush                             & -                             & -                             & -                             & -                             & -                             &-                               \\
N-ImageNet   & 2021          & Samsung-Gen3     & 1,781,167      & 1000           & 480×640             &   \XSolidBrush                            & -                             & -                             & -                             & -                             & -                             &-                               \\ \hline
DvsGesture   & 2017          & DAVIS128         & 1,342          & 11             & 128×128             & \CheckmarkBold & \XSolidBrush   & \XSolidBrush   & \XSolidBrush   & \XSolidBrush   & \XSolidBrush   & \XSolidBrush   \\
N-CARS       & 2018          & ATIS             & 24,029         & 2              & 304×240             & \CheckmarkBold & \XSolidBrush   & \XSolidBrush   & \XSolidBrush   & \XSolidBrush   & \XSolidBrush   & \XSolidBrush   \\
ASL-DVS      & 2019          & DAVIS240         & 100,800        & 24             & 240×180             & \CheckmarkBold & \XSolidBrush   & \XSolidBrush   & \XSolidBrush   & \XSolidBrush   & \XSolidBrush   & \XSolidBrush   \\
PAF         & 2019          & DAVIS346         & 450            & 10             & 346×260             & \CheckmarkBold & \XSolidBrush   & \XSolidBrush   & \XSolidBrush   & \XSolidBrush   & \XSolidBrush   & \XSolidBrush   \\
DailyAction  & 2021          & DAVIS346         & 1,440          & 12             & 346×260             & \CheckmarkBold & \XSolidBrush   & \XSolidBrush   & \XSolidBrush   & \XSolidBrush   & \XSolidBrush   & \XSolidBrush   \\
HARDVS       & 2024          & DAVIS346         & 107,646        & 300            & 346×260             & \CheckmarkBold & \CheckmarkBold   & \XSolidBrush   & \XSolidBrush   & \XSolidBrush   & \XSolidBrush   & \XSolidBrush   \\
DailyDVS-200 & 2024          & DVXplorer Lite   & 22,046         & 200            & 320×240             & \CheckmarkBold & \CheckmarkBold   & \XSolidBrush   & \XSolidBrush   & \CheckmarkBold & \CheckmarkBold & \CheckmarkBold \\
CeleX-HAR    & 2024          & CeleX-V          & 124,625        & 150            & 1280×800            & \CheckmarkBold & \XSolidBrush   & \XSolidBrush   & \XSolidBrush   & \CheckmarkBold & \CheckmarkBold & \CheckmarkBold \\ \hline
\textbf{Ours}         & 2025          & DAVIS346         & 18,041         & 62            & 346×260             & \CheckmarkBold & \CheckmarkBold & \CheckmarkBold & \CheckmarkBold & \CheckmarkBold & \CheckmarkBold & \CheckmarkBold \\ \bottomrule

\end{tabular}
}
\caption{\textit{
Comparison of event datasets for HAR. M-VW/ILL/MO, DYB, and OCC denote multiview, multi-illumination, multi-motion, dynamic background and occlusion, respectively. Note that we only report these attributes of realistic DVS datasets for HAR.
}}
\label{table1:dataset}
\vspace{-5mm}
\end{table*}


\noindent\textbf{Motion Compensation.} Event motion compensation is broadly model-driven or data-driven. Model-driven methods rely on computationally expensive optimization, such as contrast maximization \cite{gallego2017accurate, gallego2018unifying}, or joint motion estimation \cite{mitrokhin2018event, stoffregen2019event, parameshwara20210, zhou2021event}. Data-driven methods  fuse IMU data for rotational corrections \cite{delbruck2014integration, falanga2020dynamic, zhao2023event, zhou2024jstr}. In contrast, our approach avoids expensive optimization by exploiting IMU spatiotemporal correlations for real-time processing. Furthermore, we introduce a dynamic  segmentation strategy based on micro-temporal intervals, enhancing robustness over prior sensor fusion in high-speed dynamic scenarios.

\noindent\textbf{Event-Based Human Action Recognition.} Research on event camera-based action recognition primarily focuses on artificial neural networks (ANNs) and spiking neural networks (SNNs). ANN approaches \cite{Zhou_2024_CVPR, gao2023action, sabater2022event, xie2023event, wang2024event, chen2024spikmamba} have explored adaptive representations \cite{Zhou_2024_CVPR}, hypergraph networks \cite{gao2023action}, Transformers \cite{sabater2022event}, voxel embeddings \cite{xie2023event}, and emerging Mamba-based models \cite{wang2024event, chen2024spikmamba}. SNN methods \cite{cao2015spiking, bu2023optimal, ren2024spiking, guo2024ternary} leverage spike-based computation, starting from early end-to-end neuromorphic systems \cite{amir2017low}. Subsequent research has focused on bio-inspired architectures \cite{liu2021event, george2020reservoir} and training optimizations like STDP \cite{lee2018training} or deep liquid state machines \cite{soures2019deep}. SNNs are notable for their energy efficiency and temporal modeling advantages.

\label{sec:related}

\section{The DarkShake-DVS Dataset}
\label{sec:dataset}

\subsection{Image capture settings}

We constructed the DarkShake-DVS dataset, designed to serve as a benchmark for event-based HAR under low illumination with 6-DoF ego-motion. This dataset was captured using a DAVIS-346 event camera in real-world scenarios at a resolution of 346$\times$260. To explore the dynamic range capabilities of event cameras under dark conditions, we strictly adhered to recording protocols, ensuring that recording occurred in scenes with weak illumination. 

Throughout the collection process, the camera was handheld by the collector, who intentionally introduced realistic movement to simulate 6-DoF ego-motion, while IMU data including accelerometers and gyroscopes were simultaneously recorded to measure this motion. To simulate real-world environmental diversity, the dataset covers indoor and outdoor scenes, such as offices, playgrounds, kitchens, and bedrooms, with backgrounds that include both static and dynamic elements. We captured actor performances from multiple distinct viewpoints, including front, back, left, right, four diagonal directions, and upward and downward angles. In addition, we incorporated real-world situations in which hands, feet, or other objects occlude the actions.To foster further research, we will make the DarkShake-DVS dataset and our code publicly available.

To ensure action diversity, we set the duration of fine-grained actions to range from 2 to 7 seconds and recorded actions at different motion speeds, such as slow, medium, and high. The dataset comprises 62 action categories, covering complex single-person actions like splitting firewood, dancing, and cooking, as well as two-person cooperative actions like performing cardiopulmonary resuscitation. The entire data collection was completed by 15 action performers, resulting in a total of 18K video sequences. More details of the camera's optical parameters and correction operations are in the supplementary materials. Fig. \ref{fig:dataset} shows examples of actions under varying degrees of jitter, where each example contains RGB images and event frames.

\subsection{Image post-processing}

The 18K collected raw event streams underwent systematic post-processing to construct a high-quality benchmark. We manually reviewed all sequences as part of rigorous quality control and discarded invalid sequences with improperly performed actions, corrupted sensor readings, or negligible event activity, such as missed jitter or saturation under excessive illumination. We then divided the dataset into a training set (60\%), a validation set (30\%), and a test set (10\%) for standardized benchmarking. This split facilitates the fair evaluation of future event-based HAR models.

To demonstrate the challenge of DarkShake-DVS regarding 6-DoF jitter, we utilized the synchronized IMU data to eliminate subjective judgment. We calculated the average angular velocity of the gyroscope data for each sequence and used this as the classification criterion for motion intensity. Based on this criterion, our dataset can be split into three subsets with different jitter intensities: low jitter (5487 sequences, 30\%), medium jitter (7182 sequences, 40\%), and high jitter (5372 sequences, 30\%). These three subsets provide a realistic spectrum of motion intensities for evaluating the robustness of event-based HAR models.
\section{Event–IMU Stabilized HAR}
\label{sec:method}
Given the absence of existing methods that can effectively address the artifacts caused by camera motion while managing the data redundancy and non-uniform sampling challenges inherent in event streams, we propose a comprehensive benchmarking framework, which integrates an adaptive IMU-based motion compensation method (AIMC), an Iterative Greedy Sampling (IGS) module, and a Hybrid Spatio-Temporal Swin Transformer (HSTS) model.

\begin{figure*}
    \centering
  \includegraphics[scale=0.85]{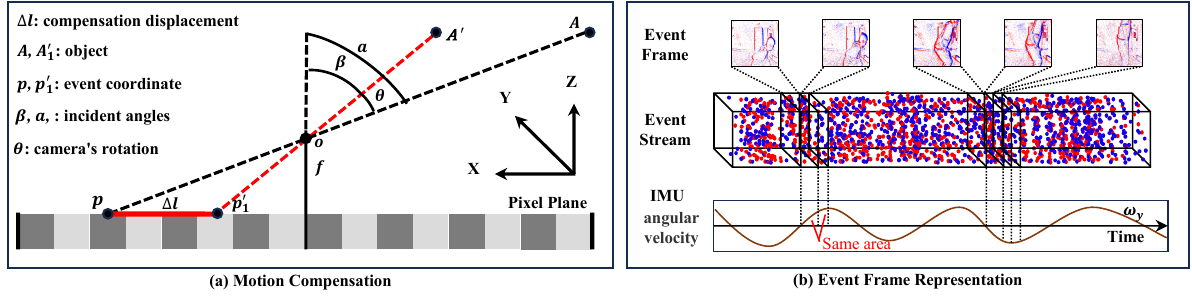}
  \captionsetup{skip=-0.1mm}
    \caption{ 
    \textit{
    \textbf{Illustration of motion compensation. } (a) gives the y-axis compensation displacement. P is the event coordinate triggered by A. After the camera rotates $\theta$ around the y-axis, $P$ moves to $P’$. The incident angles before and after the movement are $\alpha$ and $\beta$, respectively. And the compensation displacement should be the red $\Delta l$ in the pixel plane. (b) illustrates the event frame representation. In our method, the IMU data is partitioned into $N$ groups based on splitting points derived from angular velocity characteristics, including critical points, local extrema, and the median value of angular displacement within monotonic intervals. The temporal boundaries of these groups are defined by the first and last IMU timestamps within that group and are synchronized with the event stream. The compensation value for each event within a group is multiplied by a scaling factor  $\gamma$, after which all events are aggregated into a single event frame.}} 
    \label{fig:method1}
   \vspace{-5mm}
\end{figure*}

\subsection{Preliminaries}

\textbf{IMU and Event cameras.} Event cameras generate asynchronous single-pixel events ${e_i}$ when the cumulative intensity change of a specific pixel exceeds a predefined threshold. An event stream is represented as a collection of all events within a temporal window, denoted as $\varepsilon \in \mathbb{R}^{W\times H \times T} = \{ e_1,e_2,...,e_N \}$, where each event is defined by a quadruple $e_i=\{x_i,y_i,t_i,p_i \}$, $(x_i, y_i) $ denotes pixel coordinates, $t_i$ is a microsecond-level timestamp, and $p_i \in \{ +1 , -1 \}$ denotes polarity. The spatial dimensions $\{ W \times H\}$ and temporal span $T$ of the event stream collectively form the spatiotemporal representation basis for motion compensation algorithms. The DAVIS camera integrates an IMU comprising an accelerometer and a gyroscope. The IMU measures 3D linear accelerations $ \mathbf{a} = [a_x, a_y, a_z]^\top $ and angular velocities $\boldsymbol{\omega}^i = [\dot{\phi}^i, \dot{\theta}^i, \dot{\psi}^i]^\top$ in the IMU frame. Due to the extrinsic transformation between the camera and IMU frames, a rotation matrix $\mathbf{R}^{ci} \in \mathbb{R}^{3 \times 3} $ is required to align the IMU measurements with the camera frame. Specifically, angular velocities are transformed as, $\boldsymbol{\omega}^c = \mathbf{R}^{ci} \boldsymbol{\omega}^i$, where $ \boldsymbol{\omega}^c \triangleq [\dot{\phi}^c, \dot{\theta}^c, \dot{\psi}^c]^\top $ and $ \boldsymbol{\omega}^i \triangleq [\dot{\phi}^i, \dot{\theta}^i, \dot{\psi}^i]^\top  $ represent the angular velocity vectors in the camera and IMU frames. In this work, since the camera lacks depth-sensing capabilities and all sequences in our dataset were captured at approximately consistent object-camera distances, we uniformly approximate the depth as a constant value to simplify motion compensation.

\subsection{ Adaptive IMU-based Motion Compensation} 
The motion compensation method based on IMUs primarily addresses the characterization of three-dimensional rotational motion, with its focus on correcting image distortions in event cameras under intense rotational motion through Euler angles, Rodrigues' rotation formula, and quaternion methods. To address the challenge of differential displacement compensation caused by identical rotation angles in scenarios with large incidence angles, this study adopts the motion compensation framework proposed by \cite{zhao2023event}. The method calculates the total rotation angles $\phi, \theta,  \psi$ around the three axes by integrating the angular velocities over time $\delta t$, and establishes a motion compensation mapping function $\varphi: \mathbb{R}^{3} \to \mathbb{R}^{3}$ to transform the original event coordinates $(x, y, t)$ into compensated coordinates $(x^{'}, y^{'}, t)$. Specifically, the conversion between the compensated pixel coordinates $\mathbf{x}^{'}_t$ and the original coordinates $\mathbf{x}_t$ is defined as: $\mathbf{x}^{'}_t = [\mathbf{R}(\mathbf{x}_t - c_o) - \mathbf{T}] + c_o$, where $c_o \in \mathbb{R}^{2}$ denotes the center of the pixel plane, $\mathbf{R} \in \mathbb{R}^{2 \times 2}$ represents the 2D rotation matrix constructed from the rotation angle of the z-axis $\psi$, and $\mathbf{T} \in \mathbb{R}^{2}$ corresponds to the equivalent translation vector induced by rotations of the $x$-axis and the $y$-axis. 

Taking $y$-axis rotation as an example, shown in Fig. \ref{fig:method1} (a), when the initial event position lies on the $\mathbf{x}_t$-axis, the incidence angles $\alpha$ and $\beta$ between original $x_t$ and compensated $\mathbf{x}^{'}_t$ are calculated using the relationships,


\begin{equation}
\left\{
\begin{aligned}
    \alpha &= \tan^{-1}\left(\dfrac{x \cdot w}{f}\right), \\
    \beta &\approx \alpha - \theta ,
\end{aligned}
\right.
\label{eq:lif_group}
\end{equation}
where $w$ denotes pixel width and $f$ represents focal length. The nonlinear displacement compensation model, 
\begin{align}
    \Delta l &= x - \rho·\tan \beta,  \rho = f/w ,  
\end{align}
determines the $x$-axis displacement magnitude. For 3D rotational scenarios, the improved translation vector, 
\begin{align}
\mathbf{T} = (\mathbf{x}_t  - c_o) - \rho·\tan\beta ,
\end{align}
while the rotation matrix $\mathbf{R}$ maintains the standard 2D rotation form. The parameter $\beta = [\beta_x, \beta_y]^\top$ characterizes the incidence angles at compensated positions along $x$ and $y$ axes, derived by subtracting rotation angles $\theta$ and $\psi$ from original incidence angles $\alpha_x$ and $\alpha_y$ respectively. Finally, spatial mapping function $\varphi$ transforms the initial event set $\mathbf{C} \in \mathbb{R}^{3}$, generating the compensated event set $\mathbf{C}^{'} = \{x',y',t  \vert \forall(x, y, t) \in \mathbf{C}\}$, effectively eliminating spatiotemporal distortion effects caused by rotational motion.

During the implementation of the aforementioned motion compensation method, we observed that the microsecond-scale intervals between events result in compensation values on the order of $10^{-6}$ when multiplying adjacent temporal frames. However, since pixel coordinates in event stream generated frames are stored as integer types in computer systems, the integer rounding operations applied to compensated coordinates prevent positional shifts. This numerical precision limitation renders the motion compensation method physically ineffective in implementation.



 To address these limitations, we propose an adaptive motion compensation framework based on angular velocity frequency characteristics, extending the prior work in \cite{zhou2021event}. As illustrated in Fig.~\ref{fig:method1}(b), we partition the IMU angular velocity sequence into frequency-domain groups by first splitting it into positive and negative regions according to the sign (polarity) of the angular velocity,
then, within each monotonic region, we further subdivide the signal by using local extrema as splitting points, and determine the final groups based on the median cumulative angular displacement of the resulting sub-regions. The temporal boundaries of each group are defined by the first and last IMU timestamps, synchronized with event stream timestamps. 
To resolve the compensation mismatch caused by fixed scaling factors \(\gamma\) under dynamic angular velocity, we design an IMU-temporal-aware dynamic scaling mechanism, by aligning the time intervals between adjacent events \(\Delta t_{\text{event}}\) to IMU sampling intervals \(\Delta t_{\text{imu}}\), and constructing a scaling factor function \(\gamma_{\text{group}} = \gamma_{\min} + (\gamma_{\max} - \gamma_{\min}) / (a \cdot N_{\text{imu}} + b)\), where \(N_{\text{imu}}\) is the IMU sample count, \(a\) and \(b\) are tuning coefficients, and \(\gamma_{\min}\), \(\gamma_{\max}\) define the scaling bounds. Detailed ablations are provided in the supplementary material. This enables highly adaptive fine-grained compensation aligned with angular velocity dynamics.

\begin{figure*}
    \centering
  \includegraphics[scale=1]{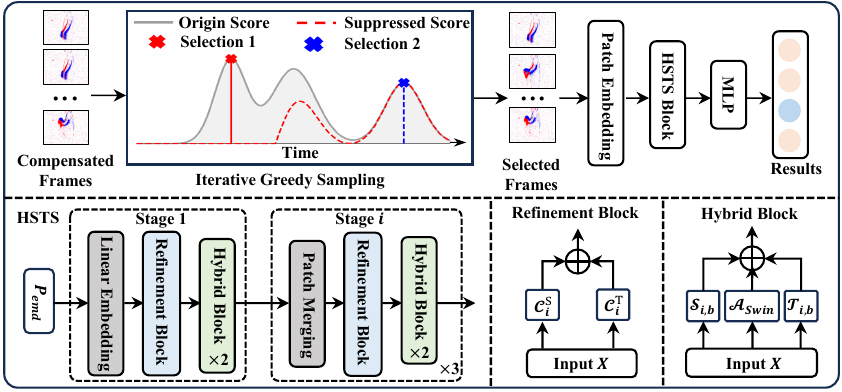}
    \caption{ 
    \textit{\textbf{Overview of our framework.} We represent event data as three-channel event images, which are processed by an Iterative Greedy Sampling (IGS) module that uses a dynamic suppression strategy to select a compact, informative set of keyframes. These keyframes are fed into the HSTS module,  which has a four-stage hybrid architecture that jointly captures long-range structure and local spatiotemporal cues. Finally, the embedding features from the HSTS module are pooled and projected to the corresponding action class.
    }}
    \label{fig:method2}
   \vspace{-4mm}
\end{figure*}
\vspace{-2mm}
\subsection{Hybrid Spatio-Temporal Swin Transformer}
Our proposed method, illustrated in Fig. \ref{fig:method2}, introduces a two-stage framework for HAR. The core of our approach consists of two sequential stages, an Iterative Greedy Sampling (IGS) module that selects a  high-quality key frame subset from the dense event stream representation, and a Hybrid Spatio-Temporal Swin Transformer (HSTS) block  which has a four-stage hybrid architecture.

\subsubsection{Iterative Greedy Sampling}
The frame sequences generated by our motion compensation method possess 
a highly variable total count, leading to significant data redundancy in long samples. Traditional uniform sampling strategies are inadequate for this task, as they often miss sparse yet crucial moments. To address this problem, we propose an adaptive iterative greedy algorithm that employs a dynamic suppression strategy to efficiently identify a maximally representative subset of keyframes.

The core idea of this algorithm is to calculate a comprehensive score $S_{\text{comb}}(i)$ for each candidate frame $i$, which is weighted by four key evaluation metrics,
\begin{align}
S_{\text{comb}}(i) = & w_{\text{rel}}{\hat{S}}_{\text{rel}}(i) + w_{\text{q}}{\hat{S}}_{\text{qual}}(i) + \nonumber \\
& w_{\text{u}}{\hat{S}}_{\text{uni}}(i) + w_{\text{d}}{\hat{S}}_{\text{div}}(i),
\end{align}
where $\hat{S}_{\text{rel}}$, which evaluates the frame's relevance to the action, and $\hat{S}_{\text{qual}}$, which evaluates the frame's quality, represent the frame's intrinsic merit; while $\hat{S}_{\text{uni}}$, which evaluates temporal uniformity, and $\hat{S}_{\text{div}}$, which evaluates visual diversity, represent dynamic suppression. The detailed parameter settings are given in the supplementary material.

As illustrated in Fig. \ref{fig:method2}, the algorithm iteratively uses this formula to build the keyframe set. In the first iteration, the method selects the highest-scoring frame based only on intrinsic merit, i.e., $\hat{S}_{\text{rel}}$ and $\hat{S}_{\text{qual}}$, which corresponds to the red X on the solid gray curve. Once this frame is selected, the dynamic suppression strategy is activated. In subsequent iterations, the $\hat{S}_{\text{uni}}$ and $\hat{S}_{\text{div}}$ scores of all remaining frames will be recalculated based on their relationship with the selected frames. Frames that are visually redundant or temporally clustered with the selected frames will have their comprehensive score $S_{\text{comb}}$ suppressed, as shown by the red dashed curve. The algorithm then selects the next highest-scoring frame from suppressed distribution, i.e., the blue X.

\subsubsection{Hybrid Spatio-Temporal Swin Transformer Block}
The $K$ sampled key frames $X \in \mathbb{R}^{C \times K \times H \times W}$ from the IGS module serve as the input to our HSTS block. As illustrated in Fig. \ref{fig:method2}, we first divide the $X$ into non-overlapping 3D patches and then project them into features $P_{emb}$,
\begin{align}
P_{emb} = \mathcal{E}_{\text{patch}}(X),
\end{align}
where $\mathcal{E}_{\text{patch}}(\cdot)$ is a 3D convolution layer. Then $P_{emb}$ are fed into the HSTS block, which contains four stages. Each stage consists of a patch merging layer, followed by a macro-level refinement block and two micro-level hybrid blocks.

\noindent\textbf{Refinement Block.} 
To mitigate semantic drift from patch merging, the input $P_{in}$	
  is first processed by a parallel refinement module before entering the hybrid blocks,
\begin{align}
P_{sp} = \mathcal{C}^{\text{S}}_i(P_{in}), P_{tp} = \mathcal{C}^{\text{T}}_i(P_{in}),
\end{align}
where, at the $i$-th stage,  $\mathcal{C}^{\text{S}}_i(\cdot)$ and $\mathcal{C}^{\text{T}}_i(\cdot)$ utilize lightweight 2D and 1D depthwise-separable convolutions, respectively, to explicitly enforce spatial consistency and temporal coherence priors. The resulting features are fused to get $P_{r}$, 
\begin{align}
P_{r} = P_{sp} + P_{tp}.
\end{align}
Then the embedding $P_{r}$ is passed into Hybrid Block.

\noindent\textbf{Hybrid Block.} The Hybrid Block is the core of our HSTS architecture. We designed three parallel paths, 
\begin{align}
P_{\mathcal{A}} = \mathcal{A}_{\text{Swin}}(P_{r}),
P_{\mathcal{L}_S} = \mathcal{L}^{\text{S}}_i(P_{r}),
P_{\mathcal{L}_T} = \mathcal{L}^{\text{T}}_i(P_{r}),
\end{align}
where the Swin-Attention \cite{liu2022video} path $\mathcal{A}_{\text{Swin}}(\cdot)$ is for modeling global correlations, the local spatial path $\mathcal{L}^{\text{S}}_i(\cdot)$ and the local temporal path $\mathcal{L}^{\text{T}}_i(\cdot)$ are used to stabilize features and suppress local noise. The outputs from the three branches are fused using learnable scalar weights $w_{\mathcal{A}}$, $w_{\mathcal{L}_S}$, and $w_{\mathcal{L}_T}$, 
\begin{align}
P_{out} = w_{\mathcal{A}} P_{\mathcal{A}} + w_{\mathcal{L}_S} P_{\mathcal{L}_S} + w_{\mathcal{L}_T} P_{\mathcal{L}_T}.
\end{align}
\noindent\textbf{Prediction.} The output embeddings $P_{out}$ from the final HSTS stage pass through a final normalization layer $\mathcal{N}(\cdot)$ and a global average pooling operator $\mathcal{P}_{\text{avg}}(\cdot)$. A final projection head $\mathcal{H}(\cdot)$ is used to predict the action logits $y$,
\begin{align}
y = \mathcal{H}(\mathcal{P}_{\text{avg}}(\mathcal{N}(P_{out}))),
\end{align}
during training, the prediction $y$ is optimized against the ground-truth class label using a cross-entropy loss.

\section{Experiment}
\label{sec:experiment}

\subsection{Dataset and Implementation details}


\begin{figure}
    \centering
  \includegraphics[scale=0.68]{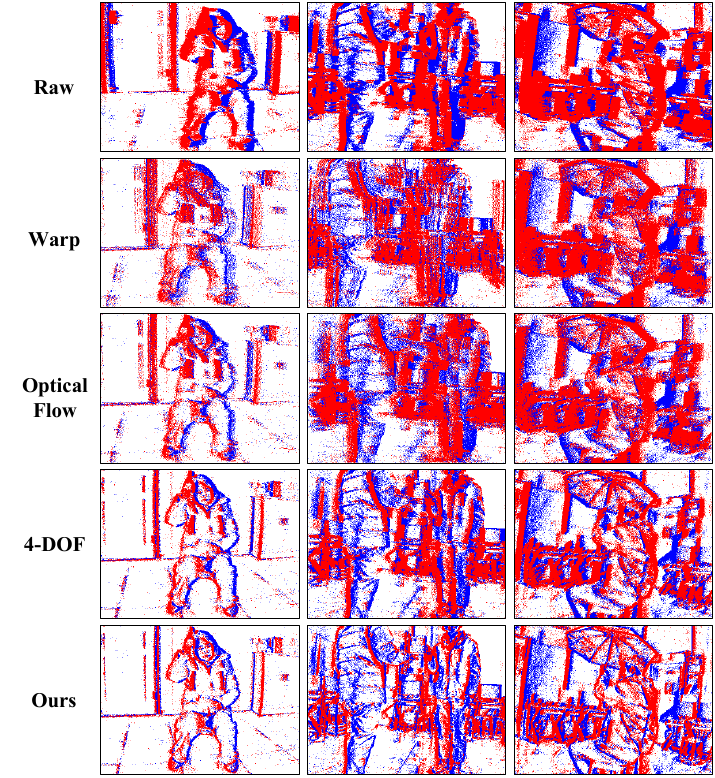}
    \caption{ 
    \textit{Motion compensation comparison with IMU-based and optimization-based methods. From the first row to the fourth row are raw event frames, compensation results from the IMU-based method \cite{zhao2023event}, and those from the optimization-based methods \cite{gallego2018unifying,mitrokhin2018event}, and ours, under challenging low-light, jitter conditions.}
    }
    \label{fig:experiment1}
    \vspace{-5mm}
\end{figure}


\noindent\textbf{Dataset.} We use three datasets to evaluate the performance of our framework, EIS-HAR. These datasets include  HARDVS \cite{wang2024hardvs}, DailyDVS-200 \cite{wang2024dailydvs} and our dataset DarkShake-DVS. Specifically, \textbf{HARDVS} \cite{wang2024hardvs}, a recently released dataset, boasts the largest number of action categories and samples, totaling 300 categories and 107,646 recordings. 
\textbf{DailyDVS-200} \cite{wang2024dailydvs} covers 200 action classes in real-world scenarios, recorded by 47 participants, and comprises 22,000 event sequences. 
\textbf{DarkShake-DVS} is described in detail above.

\noindent\textbf{Implementation details.} In training, we use a hidden embedding dimension of 96 and the Adam optimizer with a weight decay of $2e^{-2}$. The learning rate is initialized to $5e^{-4}$, and we adopt CosineAnnealingLR \cite{loshchilov2017sgdr} with a minimum learning rate of $1e^{-5}$. Our model was trained on two NVIDIA 4090 GPUs for 250 epochs with a batch size of 20.

\subsection{Motion Compensation Evaluation}

To evaluate the performance of our proposed method, we conduct comparisons on DarkShake-DVS against two categories of representative methods: an IMU-based method \cite{zhao2023event} and two optimization-based methods \cite{gallego2018unifying, mitrokhin2018event}.

We first compare our method against the IMU-based method \cite{zhao2023event} on recovering fine-grained textures from high-speed motion sequences, as shown in Fig. \ref{fig:experiment1}. Our method reconstructs visually sharper images with a higher density of pixel-level events. In contrast, \cite{zhao2023event} fails to properly handle timestamp floating-point precision, resulting in  grid-like artifacts that degrade compensation quality. Then we further compare against the optimization-based \cite{gallego2018unifying} and \cite{mitrokhin2018event}, as shown in Fig. \ref{fig:experiment1}, which rely on local optical flow and a global 4-DOF motion model, respectively. The local optical flow estimation in \cite{gallego2018unifying} lacks global constraints, causing noticeable overlapping. Although larger patches can alleviate this global inconsistency, this sacrifices local motion precision, resulting in blurred motion details.

\begin{table}[t]
\centering

\renewcommand{\aboverulesep}{0pt}
\renewcommand{\belowrulesep}{0pt}

\resizebox{\linewidth}{!}{%
\begin{tabular}{lccc}
\toprule
\textbf{Dataset} & \textbf{Algorithm} & \textbf{acc/top-1} & \textbf{acc/top-5} \\
\midrule
\multirow{11}{*}{HARDVS}
& ResNet18~\cite{he2016deep}               & 49.20 & 56.09 \\
& C3D~\cite{ji20123d}                       & 50.52 & 56.14 \\
& R2Plus1D~\cite{tran2018closer}            & 49.06 & 56.43 \\
& TSM~\cite{lin2019tsm}                     & 52.63 & 60.56 \\
& ACTION\mbox{-}Net~\cite{wang2021action}   & 46.85 & 56.19 \\
& TAM~\cite{Liu_2021_ICCV}                & 50.41 & 57.99 \\
& TimeSformer~\cite{bertasius2021space}     & 50.77 & 58.70 \\
& Swin\mbox{-}T~\cite{liu2022video}         & 51.91 & 59.11 \\
& ESTF~\cite{wang2024hardvs}                & 51.22 & 57.53 \\
& SlowFast~\cite{feichtenhofer2019slowfast} & 50.63 & 57.77 \\
\cmidrule(lr){2-4}
& \textbf{Ours}                             & \textbf{53.21} & \textbf{60.45} \\
\midrule
\multirow{12}{*}{DailyDVS-200}
& C3D~\cite{ji20123d}                       & 21.99 & 45.81 \\
& I3D~\cite{ji20123d}                        & 32.30 & 59.05 \\
& SlowFast~\cite{feichtenhofer2019slowfast}  & 41.49 & 68.19 \\
& TSM~\cite{lin2019tsm}                      & 40.87 & 71.46 \\
& EST~\cite{gehrig2019end}                   & 32.23 & 59.66 \\
& TimeSformer~\cite{bertasius2021space}      & 44.25 & 74.03 \\
& Swin\mbox{-}T~\cite{liu2022video}          & 48.06 & 74.47 \\
& ESTF~\cite{wang2024hardvs}                 & 24.68 & 50.18 \\
& GET~\cite{peng2023get}                     & 37.28 & 61.59 \\
& Spikformer~\cite{zhouspikformer}           & 36.94 & 62.37 \\
& SDT~\cite{yao2023spike}                    & 35.43 & 58.81 \\
& Evmamba~\cite{wang2024event}               & 49.65 & 75.89 \\
\cmidrule(lr){2-4}
& \textbf{Ours}                              & \textbf{51.99} & \textbf{78.13} \\
\bottomrule
\end{tabular}%
}
\caption{\textit{Overall comparison with SOTA models for event-based HAR on HARDVS and DailyDVS-200 with best results bolded.}}
\label{table1:HARDVS and daily}
\vspace{-5mm}
\end{table}

\subsection{Comparison with SOTA Methods}

Tab.~\ref{table1:HARDVS and daily}-\ref{table2:DarkShake-DVS} comprehensively compare our proposed EIS-HAR method with state-of-the-art approaches on the DarkShake-DVS, HARDVS, and DailyDVS-200 datasets. The best results are highlighted in bold. Our findings are as follows: (1) Our method outperforms all existing methods across all three datasets, achieving accuracies of 91.35\%, 53.21\%, and 51.99\%, respectively. (2) Notably, on DarkShake-DVS, SSM-based models do not perform well, and we hypothesize this is mainly due to their sensitivity to camera jitter. (3) We note that the DailyDVS-200 dataset suffers from a potential long-tail distribution; however, our method achieves an accuracy improvement through its effective extraction of spatio-temporal features. (4) As shown in Tab.~\ref{table2:DarkShake-DVS}, we present the accuracy with and without our motion compensation method, AIMC. It is observed that all methods, except for VMamba, achieve improved results after motion compensation. This indicates that our motion compensation method effectively mitigates the impact of camera motion, thereby further enhancing the models' recognition accuracy.

\subsection{Ablation Studies}

\noindent\textbf{Effectiveness of Motion Compensation.} To validate the necessity of our motion compensation module, we compare the model's performance on the original event stream versus the compensated event stream. As shown in Tab.~\ref{table4:Ablation}, motion compensation improves the accuracy by 2.74\%. Furthermore, the qualitative results in Fig.~\ref{fig:experiment1} demonstrate that the compensated event frames effectively eliminate motion blur and reconstruct sharper object contours. These results prove that motion compensation is a critical preprocessing step.

\noindent\textbf{Effectiveness of Iterative Greedy Sampling.} 
Tab.~\ref{table4:Ablation} shows that replacing IGS with uniform sampling reduces accuracy from 91.35\% to 85.76\%, demonstrating the effectiveness of IGS.
Uniform sampling inevitably selects frames with low information density or captures multiple consecutive frames with high similarity, leading to information redundancy. In contrast, our strategy adaptively selects keyframes rich in dynamic information, leading to more informative and diverse key frames for event-based HAR.

\noindent\textbf{Effectiveness of the HSTS Architecture.} We compare our full HSTS with two ablated variants: w/o Re, which removes the Refinement Block; and w/o Hi, which removes the spatial and temporal paths of the Hybrid Block. As shown in Tab.~\ref{table4:Ablation}, our complete model consistently outperforms both variants. Removing Re weakens feature refinement, whereas removing Hi discards complementary spatial–temporal cues and harms long-range modeling. These results validate the effectiveness of our dual-stage design.

\begin{table}[t]
\centering
\small

\renewcommand{\aboverulesep}{0pt}
\renewcommand{\belowrulesep}{0pt}

\begin{tabularx}{0.47\textwidth}{lYYYY}
\toprule
\textbf{Algorithm} & \textbf{Publish} & \textbf{\#P} & \textbf{\textit{w/o} AIMC} & \textbf{\textit{w/} AIMC} \\
\midrule
ResNet50 \cite{he2016deep}                 & CVPR$'$16 & 11.7M  & 82.77 & 84.34 \\
TSM \cite{lin2019tsm}                      & ICCV$'$19 & 24.3M  & 81.41 & 83.55 \\
SlowFast \cite{feichtenhofer2019slowfast}  & ICCV$'$19 & 33.6M  & 83.91 & 87.25 \\
TAM \cite{Liu_2021_ICCV}                 & ICCV$'$21 & 25.6M  & 79.60 & 84.12 \\
TimeSformer \cite{bertasius2021space}      & ICML$'$21 & 121.2M & 81.74 & 85.63 \\
Swin\mbox{-}T \cite{liu2022video}          & CVPR$'$22 & 27.8M  & 87.24 & 88.86 \\
Spikformer \cite{zhouspikformer}           & ICLR$'$23 & 23.4M  & 80.17 & 85.77 \\
ESTF \cite{wang2024hardvs}                 & AAAI$'$24 & 46.7M  & 84.64 & 87.63 \\
Vision Mamba \cite{zhu2024vision}          & ICML$'$24 & 26.0M  & 65.89 & 66.77 \\
VMamba \cite{liu2024vmamba}                & NIPS$'$24 & 76.5M  & 67.73 & 65.11 \\
VideoMamba \cite{li2024videomamba}         & ECCV$'$24 & 26.0M  & 60.55 & 61.44 \\
\midrule
\textbf{Ours}                              & --        & 34.0M  & \textbf{88.61} & \textbf{91.35} \\
\bottomrule
\end{tabularx}
\caption{\textit{Experimental results on the DarkShake-DVS dataset. We present the recognition accuracy with and without AIMC under identical settings, with the highest values highlighted in bold.
}}
\label{table2:DarkShake-DVS}
\vspace{-2mm}
\end{table}

\begin{figure}
    \centering
  \includegraphics[scale=0.7]{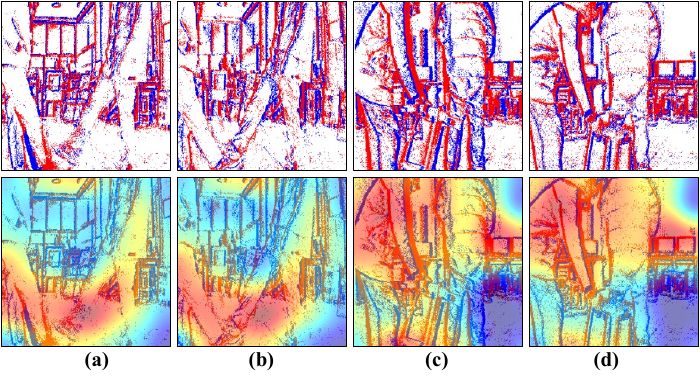}
  \captionsetup{skip=-0.1mm}
    \caption{ 
    \textit{Attention maps  of HSTS. High-attention regions are marked in red, and low-attention regions are marked in blue.} 
    }
    \label{fig:experiment2}
   \vspace{-5mm}
\end{figure}

\subsection{Discussion}

\noindent\textbf{Computation Times.}
As shown in Tab.~\ref{tablelast:time}, on a single thread of an AMD EPYC 7B12 64-Core Processor, \cite{mitrokhin2018event} requires 10 ms per iteration, needing over 30 steps to converge, and \cite{gallego2018unifying} needs 210 ms to converge. In contrast, our method completes the entire compensation process in 70ms. This efficiency, combined with a significantly higher pixel-event density, demonstrates that our IMU-based compensation achieves a substantial reduction in computational cost.

\begin{figure}
    \centering
  \includegraphics[scale=0.7]{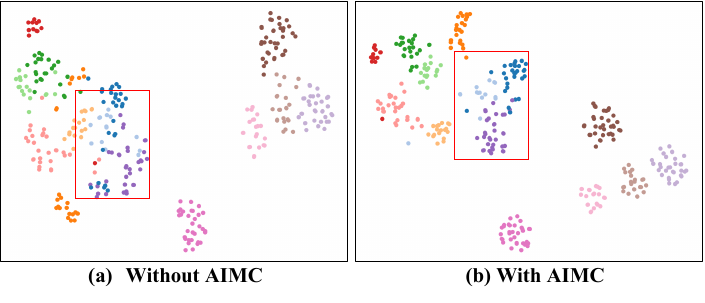}
  \captionsetup{skip=-0.1mm}
    \caption{ 
    \textit{Visualization of the feature distribution for 15 action classes in DarkShake-DVS using t-SNE contrasts panel (a) without AIMC and panel (b) with AIMC, with red boxes highlighting groups of classes exhibiting highly similar feature representations. }}
    \label{fig:tsen}
\vspace{-5mm}
\end{figure}

\noindent\textbf{Visualization.} As shown in Fig.~\ref{fig:tsen}, we visualize features from 15 DarkShake-DVS action classes using t-SNE. We observe that without motion compensation, the discriminability of certain samples is low, as highlighted by the red bounding box. Crucially, even after compensation, some categories remain indistinguishable. We hypothesize this is due to the excessive magnitude of jitter within these specific dataset categories, leading to feature convergence.

\begin{table}[t]
\centering
\small

\renewcommand{\aboverulesep}{0pt}
\renewcommand{\belowrulesep}{0pt}

\begin{tabularx}{0.47\textwidth}{lYYYYY}
\toprule
\textbf{Case} & \textbf{AIMC} & \textbf{IGS} & \textbf{Re} & \textbf{Hi} & \textbf{Results} \\
\midrule
\textit{w/o} AIMC &  & \CheckmarkBold & \CheckmarkBold & \CheckmarkBold & 88.61 \\
\textit{w/o} IGS  & \CheckmarkBold &  & \CheckmarkBold & \CheckmarkBold & 85.76 \\
\textit{w/o} Re   & \CheckmarkBold & \CheckmarkBold &  & \CheckmarkBold & 89.43 \\
\textit{w/o} Hi   & \CheckmarkBold & \CheckmarkBold & \CheckmarkBold &  & 87.36 \\
\midrule
\textbf{Ours}     & \CheckmarkBold & \CheckmarkBold & \CheckmarkBold & \CheckmarkBold & \textbf{91.35} \\
\bottomrule
\end{tabularx}
\caption{\textit{Ablation study of AIMC, IGS, Re, and Hi on the DarkShake-DVS. w/o IGS replaces our IGS module with uniform sampling. Re denotes the Refinement Block in HSTS, and Hi denotes the spatial and temporal paths in the Hybrid Block.}}
\label{table4:Ablation}
\vspace{-2mm}
\end{table}





\begin{table}[t]
\centering
\small
\renewcommand{\aboverulesep}{0pt}
\renewcommand{\belowrulesep}{0pt}

\begin{tabularx}{0.47\textwidth}{lYYYY}
\toprule
\textbf{Method} & \textbf{Time} & \textbf{Yaw} & \textbf{Pitch} & \textbf{Roll} \\
\midrule
Optical Flow \cite{gallego2018unifying} & 210ms & 3.21 & 2.22 & 1.74 \\
4-DOF \cite{mitrokhin2018event}                  & 300ms & 3.42 & 2.12 & 1.82 \\
\textbf{Ours}                         & \textbf{70ms} & \textbf{3.74} & \textbf{2.45} & \textbf{2.13} \\
\bottomrule
\end{tabularx}
\caption{
\textit{Comparison of computation time and pixel-event density for yaw, pitch , and roll between our method and \cite{gallego2018unifying, mitrokhin2018event}. }
}
\label{tablelast:time}
\vspace{-2mm}
\end{table}




\section{Conclusion}
\label{sec:conclusion}


In this work, we address the performance challenges of event-based HAR under low-light and severe jitter. We introduce the DarkShake-DVS dataset, the first large-scale HAR benchmark featuring authentic 6-DoF motion and synchronized IMU data. Furthermore, we design an IMU-based adaptive motion compensation framework that efficiently removes spatio-temporal distortions and substantially outperforms optimization methods, completing compensation in just 70ms. Our complete framework establishes a SOTA performance across all tested datasets, laying the foundation for future robust event-based HAR systems.

\section{Acknowledgment}

This work is supported by the National Natural Science Foundation of China 62302045, the Fundamental Research Funds for the Central Universities, and the Beijing Institute of Technology BIT Special-Zone.
{
    \small
    \bibliographystyle{ieeenat_fullname}
    \bibliography{main}
}

\end{document}